\documentclass[conference]{IEEEtran}
\IEEEoverridecommandlockouts
\usepackage{cite}
\usepackage{amsmath,amssymb,amsfonts}
\usepackage{algorithmic}
\usepackage{graphicx}
\usepackage{textcomp}
\usepackage{xcolor}
\usepackage{cite}
\usepackage{authblk}

\def\BibTeX{{\rm B\kern-.05em{\sc i\kern-.025em b}\kern-.08em
    T\kern-.1667em\lower.7ex\hbox{E}\kern-.125emX}}
\begin{document}

\title{Self-Supervised Joint Learning Framework of Depth Estimation via Implicit Cues}

\author[1]{Jianrong Wang}
\author[1]{Ge Zhang}
\author[1]{Zhenyu Wu}
\author[1]{XueWei Li\thanks{* Corresponding author}}
\author[2,*]{Li Liu}

\affil[1]{College of Intelligence and Computing, Tianjin University}
\affil[2]{Shenzhen Research Institute of Big Data, the Chinese University of Hong Kong, Shenzhen \authorcr Email: \{wjr, zzgg, francis\_wu, lixuewei\}@tju.edu.cn, liuli@cuhk.edu.cn}

\maketitle

\begin{abstract}
In self-supervised monocular depth estimation, the depth discontinuity and motion objects' artifacts are still challenging problems. Existing self-supervised methods usually utilize a single view to train the depth estimation network. Compared with static views, abundant dynamic properties between video frames are beneficial to refined depth estimation, especially for dynamic objects. In this work, we propose a novel self-supervised joint learning framework for depth estimation using consecutive frames from monocular and stereo videos. The main idea is using an implicit depth cue extractor which leverages dynamic and static cues to generate useful depth proposals. These cues can predict distinguishable motion contours and geometric scene structures. Furthermore, a new high-dimensional attention module is introduced to extract clear global transformation, which effectively suppresses uncertainty of local descriptors in high-dimensional space, resulting in a more reliable optimization in learning framework. Experiments demonstrate that the proposed framework outperforms the state-of-the-art (SOTA) on KITTI and Make3D datasets.
\end{abstract}

\section{Introduction}

Depth and ego-motion estimations play essential roles in understanding geometric scenes from videos and images, and have broad applications such as robotics~\cite{982903} and autonomous driving~\cite{7410669}. Supervised models~\cite{Hu2018Revisiting,8578312,7346484,DBLP:journals/corr/abs-1908-03706} have obtained depth maps with vibrant details from color images. However, it is difficult and expensive to accurately collect large-scale labels in practice, and these supervised models are only suitable for specific scenarios. 

In recent years, self-supervised methods have attracted increasing interests, and there have been some successes ~\cite{8100183,DBLP:conf/cvpr/MahjourianWA18,8578310,DBLP:conf/eccv/YangWWXN18,DBLP:journals/corr/abs-1806-01260,DBLP:conf/nips/BianLWZSC019}. In the absence of ground truth, one can still recover scene depth and ego-motion from monocular video sequences using self-supervised methods. The key idea is that one can first warp the source view to the target view through the estimated depth of scenes and ego-motion of the camera, and then simultaneously optimize the \textit{depth estimation network} (DepthNet) and the \textit{pose estimation network} (PoseNet) by minimizing the view reconstruction loss.

\begin{figure*}
\centering
\includegraphics[width = 14cm]{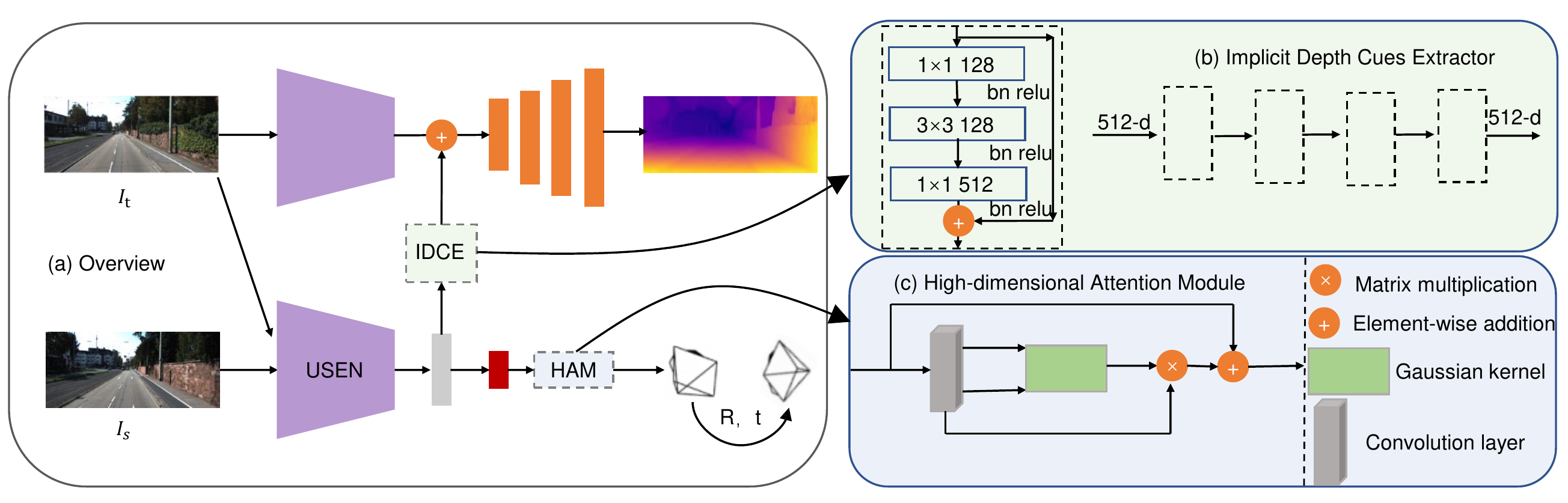}
\caption{Outline of the proposed joint learning framework. (a) Overview of the framework: a DepthNet for depth estimation and a PoseNet that takes two stacked input frames to estimate the relative camera pose. The PoseNet encoder is a \textit{unit stream extraction network} (USEN). (b) and (c) are the detailed structures of the two proposed modules, i.e., IDCE and HAM, respectively. IDCE extracts static and dynamic depth cues by cascading four identical bottlenecks, and HAM extracts global dynamic transformation from the unit stream using convolutions and Gaussian kernel. IDCE connects the USEN and the DepthNet decoder. For training, source view $I_s \in \{ I_{t-1},I_{t+1},I_t^{\textrm{stereo}}\}$, and IDCE is valid for $I_{t-1}$, while for evaluation, $I_s \in \{ I_{t-1},I_{t}\}$. }
\label{fig:overview}
\end{figure*}

 However, this framework has the following deficiencies. (1) The DepthNet only uses the static information of the current view and does not effectively utilize the rich dynamic and static depth cues between adjacent views. Hence, the predicted depth sizes of adjacent frames are inconsistent, and the depth between frames shows a leap of change. (2) The existing masking methods~\cite{8578310,8953778,DBLP:conf/eccv/ZouLH18} filter out pixels where there is object motion in the scene, so failures on motion scenes are not penalized enough. More precisely, the network comes to a deadlock and cannot seek the global optimal solution, which results in artifacts of moving objects.

The main goal of this work is to effectively alleviate the problems of depth discontinuity and motion objects' artifacts as mentioned above. We take advantage of the temporal and spatial information changes between consecutive frames, which we call \emph{unit stream} in this work. Based on the unit stream and the mainstream framework adopted in ~\cite{8100183, DBLP:conf/nips/BianLWZSC019, DBLP:conf/cvpr/MahjourianWA18} consisting of the DepthNet and PoseNet, we propose a novel self-supervised joint learning framework as shown in Fig. \ref{fig:overview}. This framework introduces two efficient modules to utilize unit stream. (1) The first module \textit{implicit depth cues extractor} (IDCE) connects the DepthNet and PoseNet. IDCE automatically selects reliable cues to constrain static and dynamic geometric scenes. 
The unit stream is modeled via statistics of convolutional activations to extract implicit dynamic/static cues and produce powerful depth proposals. The proposals are able to guide subsequent scene depth and make depth estimation near dynamic objects more accurate, while static cues enforce the DepthNet to predict smooth depth changes over consecutive snippets. (2) The second module \textit{high-dimensional attention module} (HAM) obtains more robust camera pose for accurate view reconstruction. It extracts global dynamic transformation from the unit stream by using convolutions and Gaussian kernel. This module effectively suppresses the uncertainty of the local descriptor in the high-dimensional space and coordinates the depth network to learn better weights. Note that the proposed framework and modules can be generalized to other existing self-supervised depth estimation methods.

To summarize, our main contributions are three-fold:
\begin{itemize}
\item[$\bullet$] To the best of our knowledge, this is the first work that propose a novel module called IDCE to connect the DepthNet and PoseNet in order to extract implicit static and dynamic depth cues from the shallow space of the unit stream.
\item[$\bullet$] The novel HAM captures global pose transformation from the unit stream, making the joint learning framework optimization more efficient. Besides, it can be used as a post-processing method for other pose estimation networks.
\item[$\bullet$] The joint learning framework is extensively evaluated on KITTI~\cite{DBLP:journals/ijrr/GeigerLSU13} and  Make3D~\cite{4531745} datasets. Experimental results show that the proposed framework achieves SOTA performance, and outperforms most of recent algorithms by a significant margin.
\end{itemize}

\section{Related Work}

Depth estimation has been studied for a long time. In this section, we mainly discuss the related works based on deep learning from two perspectives: supervised and self-supervised depth estimation.

\subsection{Supervised depth estimation}
In recent years, deep learning has made a breakthrough in depth estimation. Supervised depth estimation~\cite{7346484} seeks a mapping from color images to depth maps.
Eigen et al.\cite{DBLP:conf/nips/EigenPF14} first employed a multi-scale convolutional neural network, which refined the estimated depth map from low spatial resolution to high spatial resolution. In order to overcome the low-resolution problem, Laina et al.\cite{7785097} employed an up-sampling method for learning. Fu et al.\cite{8578312} introduced a spacing-increasing discretization strategy to discretize depth, and then adopted a multi-scale dilated convolution to capture multi-scale information in parallel. 

Although these supervised methods have achieved excellent performance, they need ground truth labels collected by expensive LIDAR~\cite{DBLP:journals/ijrr/GeigerLSU13} or RGBD cameras\cite{DBLP:conf/eccv/SilbermanHKF12}, which place restrictions on usage scenarios or depth ranges.

\subsection{Self-supervised depth estimation}
Without requiring the ground truth labels, self-supervised methods use photometric constraints from multiple views, e.g., multiple views captured by a monocular camera or stereo~\cite{DBLP:journals/corr/abs-1907-05820,DBLP:conf/eccv/GuoLYRW18,DBLP:conf/icra/LiWLG18,DBLP:conf/cvpr/MahjourianWA18,8100183}. The following discussions mainly focus on these two aspects.

\subsubsection{Stereo depth estimation}
Garg et al.~\cite{DBLP:conf/eccv/GargKC016} leveraged the epipolar geometry~\cite{DBLP:books/daglib/0015576} inherent in stereoviews to train the monocular DepthNet, where the photometric consistency loss between stereo pairs is used as the supervision signal. Godard et al.~\cite{8100182} proposed a left-right consistency constraint between left and right disparity maps. In these methods, accurately rectifying stereo cameras provide explicit pose supervision for self-supervised depth estimation.

\subsubsection{Monocular depth estimation}
SfMLearner~\cite{8100183} was the first method to learn both depth and ego-motion using the geometric constraints of monocular video. Meanwhile, additional masks ignored moving objects that violated the rigid scene assumption. Following this framework, some approaches in~\cite{8578310,8953778,DBLP:journals/corr/abs-1806-01260,DBLP:conf/nips/BianLWZSC019,DBLP:conf/eccv/ZouLH18} have been proposed to solve the challenge of moving objects. Although they show significant improvements in the performance, they still suffer from ineffective issues from dynamic scenes in a monocular setting. These methods pay little attention to moving areas or discard them directly, and thus the network comes to a deadlock and cannot calculate the global minima in areas with motion. As a result, the DepthNet cannot predict distinguishable motion contours and geometric scene structure. Casser et al.~\cite{DBLP:conf/aaai/CasserPMA19} proposed a novel approach that modelled moving objects and produced higher quality results. Besides, it was proposed in~\cite{DBLP:conf/cvpr/AbarghoueiB18,bozorgtabar2019syndemo} that utilizing synthetic data can collect diverse training data. Yang et al.~\cite{8578129} combined the normal and edge geometry to achieve better performance. Very recently, Patil et al.~\cite{DBLP:journals/corr/abs-2001-02613} exploited the recurrent neural network (RNN) to generate a time series of depth maps. Although they use spatio-temporal information, the complex network structure creats huge computational costs during training.

Unlike the works mentioned above, based on the general framework, we propose a new self-supervised joint learning framework that connects the DepthNet and PoseNet to extract implicit static and dynamic depth cues from the shallow space of the unit stream. Moreover, the global dynamic transformation from the unit stream is also exploited.

\section{Method}

In this section, we mainly introduce the proposed joint learning framework, which takes the adjacent video frames $I_{t-1}$, $I_t$ as input, and a depth map $D_t$ as output. Details of the two proposed modules, IDCE and HAM, will be described. Before that, we first review the key ideas of the commonly used baseline in self-supervised depth estimation.

\subsection{Algorithm baseline}
The baseline consists of two networks, i.e., the DepthNet and the PoseNet. The former one aims to estimate the dense depth map of the target view, and the latter aims to estimate the relative camera pose between nearby views for monocular and mixed (i.e., monocular and stereo) training. In the absence of ground truth, the DepthNet and PoseNet can be solely optimized using the view reconstruction loss between the original target view and the synthesized target view.

According to~\cite{8100183}, the view $I_t'$ can be synthesized from $I_s$ as:
\begin{equation}
p_{s\rightarrow t} \sim KT_{t\rightarrow s}D_t \left(p_t \right)K^{-1}p_t,
\end{equation}
where $D_t$ is the predicted depth of target view $I_t$, $T_{t\rightarrow s}$ is the relative camera pose of the source view $I_s$  with respect to the target view $I_t$, $p_t$ and $p_{s\rightarrow t}$ are the homogeneous coordinates of a pixel in $I_t$ and $I_t'$, respectively. $K$ is the camera intrinsic matrix. During training process of the self-supervised model with stereoviews, $D_t$ is the only unknown variable. However, for monocular training, the source view is part of the temporally adjacent frames $\left(I_s \in \{ I_{t-1},I_{t+1}\}\right)$, and thus the relative camera pose $T_{t\rightarrow s}$ also needs to be predicted.
For mixed training, the source view $I_s$ is part of temporally adjacent frames and the opposite stereo view  $\left(I_s \in \{ I_{t-1},I_{t+1},I_t^{\textrm{stereo}}\}\right)$.

Concerning the loss function, following~\cite{DBLP:journals/corr/abs-1806-01260}, a common total loss is composed of photometric loss and smoothness loss:
\begin{align}
L_{\textrm{total}} &= \sum \limits_{s} \mu L_{\textrm{ph}}^s + \gamma L_{\textrm{smooth}}^{s},
\label{equ:total_loss} \\
\textrm{with} \ \ \ \mu &= [ \mathop{\min}\limits_{I_s} L_{\textrm{ph}} (I_t, I_t') < \mathop{\min}\limits_{I_s} L_{\textrm{ph}}(I_t, I_s)],
\end{align}
where $\mu$ is the auto-masking loss, s is the scale index value, and $\gamma$ is a hyperparameter, which is set to be 0.001. The average loss at multiple scales is taken as the final loss. 

In equation~\eqref{equ:total_loss}, the photometric loss $L_{\textrm{ph}}$ shown in equation~\eqref{equ:final_photometric_loss} is a combination of the structural similarities (SSIM)~\cite{1284395} and $L_1$ loss for multiple reconstructed views.
\begin{equation}
L_{\textrm{ph}}(I_t,I_t') = \sum_{p}\mathop{\min}\limits_{I_s} ( \alpha L_{\textrm{ssim}}+(1 - \alpha) \|I_t \left(p\right)-I_t'\left(p\right)\|_1  ),
\label{equ:final_photometric_loss}
\end{equation}
where $p$ is the index value of pixel coordinates. $\alpha$ is a hyper-parameter that set to be 0.85 and $L_{\textrm{ssim}}$ denotes:
\begin{equation}
L_{\textrm{ssim}} = \frac{1-\textrm{SSIM}(I_t,I_t')}{2}.
\label{equ:ssim}
\end{equation}

Here, the per-pixel minimum reprojection loss is adopted to calculate the minimum photometric error at various scales of all source views.

Besides, in equation~\eqref{equ:total_loss}, the edge-aware depth smoothness loss $L_{\textrm{smooth}}$ in~\cite{8100182} is also employed.
\begin{equation}
L_{\textrm{smooth}}=\sum \limits_{i,j} |\partial_{x} D_t^{i,j}| e^{-| \partial_{x} I_t^{i,j}|}+  |\partial_{y} D_t^{i,j}| e^{-| \partial_{y} I_t^{i,j}|},
\end{equation}
where $D_t^{i,j}$ is the mean-normalized inverse depth map. $i, j$ denote pixel index value of $I_t$. $\partial_{x}$ and $\partial_{y}$ denote gradients in the $x$ and $y$ directions, respectively. Applying such regularization enforces the DepthNet to produce sharp edge distribution at sharply varying pixels while producing smooth depth in continuous regions.

\begin{table*}
	\centering
	\caption{Evaluation results of depth estimation on the KITTI test set~\cite{DBLP:conf/nips/EigenPF14}. The methods trained on KITTI raw dataset~\cite{DBLP:journals/ijrr/GeigerLSU13} are denoted by K, virtual KITTI dataset  are denoted by vK, and models with pre-training on CityScapes~\cite{7780719} are denoted by CS+K. M, S and D$^{*}$ denotes monocular video, stereo supervision and auxiliary depth supervision, respectively. D means depth supervision. The best results in each category are in bold, and the second best are underlined.}

		\begin{tabular}{|c c|c c c c|c c c|}
			\hline
			
			Methods & Dataset & Abs Rel & Sq Rel &RMSE & RMSE log & {$\delta$ $ \textless$ 1.25 } & {$\delta$ $ \textless$ $1.25^2$} & {$\delta$ $ \textless$ $1.25^3$}\\
			\hline
			\hline
			
			Eigen et al.~\cite{DBLP:conf/nips/EigenPF14}   &K (D)		&0.203 &1.548 &6.307 &0.282& 0.702 &0.890 & 0.958\\
			Liu et al.~\cite{7346484}     &K (D) 		&0.202 &1.614 &6.523 &0.275& 0.678 &0.895 & 0.965\\
			Garg et al.~\cite{DBLP:conf/eccv/GargKC016}& K (S)			&0.152& 1.226& 5.849& 0.246& 0.784& 0.921& 0.967\\
			Godard et al.~\cite{8100182} &CS+K (S)		&0.124 &1.076 &5.311 &0.219& 0.847 &0.942 & 0.973\\
			
			Yang et al.~\cite{DBLP:conf/eccv/YangWWXN18} &K+CS (S)		&0.114 &1.074 &5.836 &0.208& 0.856 &0.939 & 0.976\\
			Guo et al.~\cite{DBLP:conf/eccv/GuoLYRW18} & K (DS) 				&\underline{0.096}& \underline{0.641}&\underline {4.095}&\underline{0.168}&\underline{0.892}& \underline{0.967}& \underline{0.986}\\
			DORN~\cite{8578312} & K (D) 				&\textbf {0.072} &\textbf{0.307} &\textbf{2.727} &\textbf{0.120}&\textbf{0.932} &\textbf{0.984} & \textbf{0.994}\\	
			
			\hline
			Zhou et al.~\cite{8100183} &K (M) 			&0.208 &1.768& 6.856 &0.283 &0.678& 0.885& 0.957\\
			Yang et al.~\cite{DBLP:conf/aaai/YangWXZN18} &K (M)			&0.182& 1.481& 6.501& 0.267& 0.725 &0.906& 0.963\\
			Mahjourian et al.~\cite{DBLP:conf/cvpr/MahjourianWA18} &K (M)	&0.163& 1.240& 6.220& 0.250& 0.762& 0.916 &0.968\\
			Wang et al.~ \cite{8578314} & K (M) 		&0.151& 1.257& 5.583& 0.228& 0.810& 0.936& 0.974\\
			GeoNet~\cite{8578310} &K (M)				&0.149 &1.060& 5.567 &0.226& 0.796& 0.935 &0.975\\
			DF-Net~\cite{DBLP:conf/eccv/ZouLH18} & K (M)				& 0.150& 1.124 &5.507& 0.223 &0.806& 0.933 &0.973\\
			Ranjan et al.~\cite{8953778} & K (M)		& 0.140& 1.070& 5.326& 0.217& 0.826& 0.941& 0.975\\
			Struct2depth~\cite{DBLP:conf/aaai/CasserPMA19} &K (M) & 0.141& 1.026& 5.291 &0.215& 0.816& 0.945& 0.979\\
			SynDeMo~\cite{bozorgtabar2019syndemo}  &K+vK (MD$^*$)& 0.116 &\textbf{0.746}  & \textbf{4.627} & 0.194 & 0.858  & 0.952 & 0.977\\
			GLNet~\cite{DBLP:journals/corr/abs-1907-05820} &  K (M)				& 0.135& 1.070& 5.230& 0.210& 0.841& 0.948& 0.980\\
			Zhou et al.~\cite{DBLP:journals/corr/abs-1910-08897} (384$\times$1248)&K (M)	& 0.121& 0.837& 4.945& 0.197& 0.853& 0.955& \textbf{0.982}\\
			Monodepth2~\cite{DBLP:journals/corr/abs-1806-01260} & K (M)			&{ 0.115}& 0.903& 4.863& \underline{0.193}&{ 0.877}& \underline{0.959}& \underline{0.981}\\
			Bian et al.~\cite{DBLP:conf/nips/BianLWZSC019} & K (M)			& 0.137& 1.089& 5.439& 0.217& 0.830& 0.942& 0.975\\
			Patil et al.~\cite{DBLP:journals/corr/abs-2001-02613} & K (M)			& \underline{0.111}& 0.821& \underline{4.650}&\textbf {0.187}&\underline{0.883} &\textbf {0.961}& \textbf {0.982}\\
			
			\textbf {Ours} (192$\times$640)& K (M)	&\textbf{ 0.106}& \underline{0.799}& {4.662}&\textbf {0.187}& \textbf{0.889}&\textbf{0.961}&\textbf {0.982}\\
			\hline 
			\textbf {Ours} (320$\times$1024)& K (M)				&\textbf {0.106}&\underline {0.773}&\textbf{4.491}& \textbf{0.185}& \textbf{0.890}&\textbf {0.962}&\textbf{ 0.982}\\
			\hline
			
			Monodepth2~\cite{DBLP:journals/corr/abs-1806-01260} (192$\times$ 640)& K (MS)		& 0.106& 0.818& 4.750& 0.196& 0.874& 0.957& 0.979\\
			Watson et al.~\cite{watson2019self} (192$\times$ 640)& K (MSD$^*$) & 0.106& 0.780& 4.695& 0.193& 0.875& 0.958& 0.980\\
			\textbf {Ours} (192$\times$640)& K (MS)				& \textbf{0.102}& \textbf{0.776}&\textbf{ 4.534}&\textbf{0.183}& \textbf{0.893}& \textbf{0.963}& \textbf{0.982}\\
			\hline
			
			Monodepth2~\cite{DBLP:journals/corr/abs-1806-01260} (320$\times$1024)&K (MS)		& 0.106& 0.806& 4.630& 0.193& 0.876& 0.958& 0.980\\
			Watson et al.~\cite{watson2019self} (320$\times$1024) & K (MSD$^*$)& \textbf{0.100}& 0.728& 4.469& 0.185& 0.885& 0.962& 0.982\\
			\textbf {Ours}(320$\times$1024)& K (MS)			&\underline{0.101}&\textbf{ 0.725}&\textbf{ 4.360}& \textbf{0.179}&\textbf{ 0.898}& \textbf{0.965}& \textbf{0.983} \\ 
			\hline
			GLNet~\cite{DBLP:journals/corr/abs-1907-05820}    & CS+K (M)		& 0.129& 1.044& 5.361& 0.212& 0.843& 0.938& 0.976\\
			Bian et al.~\cite{DBLP:conf/nips/BianLWZSC019} & CS+K (M)		& 0.128& 1.047& 5.234& 0.208& 0.846& 0.947& 0.976\\
			Struct2depth~\cite{DBLP:conf/aaai/CasserPMA19}  & CS+K (M) 	& 0.108& 0.825& 4.750& 0.186& 0.873& 0.957& 0.982\\
			SynDeMo~\cite{bozorgtabar2019syndemo} & CS+K+vK (MD$^*$)	& 0.112& \textbf{0.740}& \textbf{4.619}& 0.187& 0.863& 0.958& 0.983\\
			\textbf {Ours}(192$\times$640)& CS+K (M)				& \textbf{0.106}&  \underline{0.774}& \underline{4.623}& \textbf{0.184}& \textbf{0.886}& \textbf{0.962}& \textbf{0.983} \\
			
			\hline
	\end{tabular}
	
	\label{table:T1}
\end{table*}

\begin{figure*}
	\centering
	\includegraphics[width=16cm]{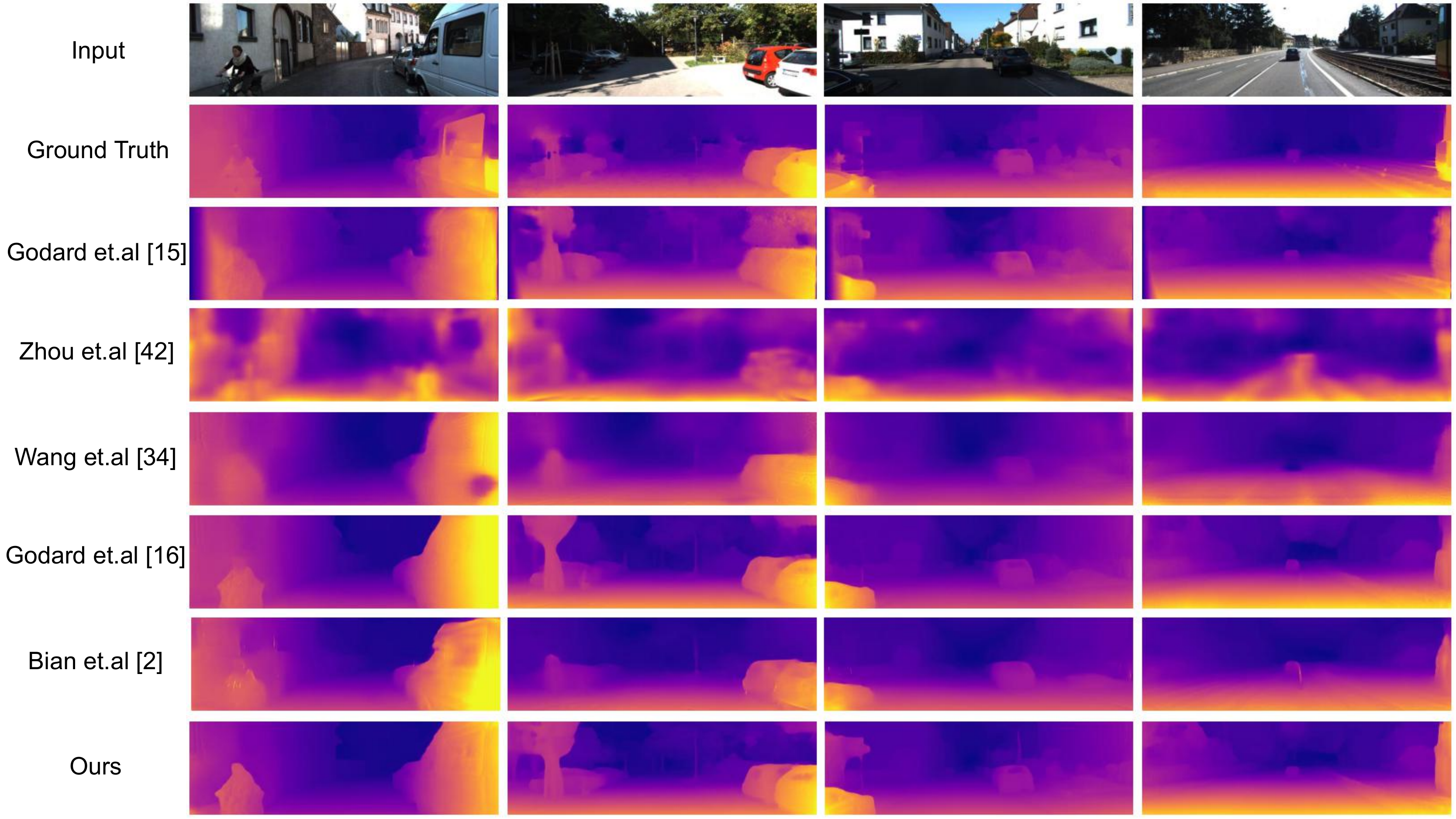}
	\caption{ Qualitative results on the KITTI Eigen split. 
		From top to bottom, the images are input, ground truth, results of Godard et al.~\cite{8100182}, Zhou et al.~\cite{8100183}, Wang et al.~\cite{8578314}, Godard et al.~\cite{DBLP:journals/corr/abs-1806-01260}, Bian et al.~\cite{DBLP:conf/nips/BianLWZSC019} and our KITTI monocular method, respectively. Our method effectively solves the motion blur and artifact, provides a clearer motion contour, and offers sharper predictions of static objects.}
	\label{fig:results}
\end{figure*}
\subsection{Self-supervised joint learning framework}
\subsubsection{Motivation}
The mainstream framework only considers the camera pose information in the unit stream, and ignores the role of depth cues between adjacent frames. Motivated by this, we consider both implicit depth cues and pose information to be important attributes of unit stream, and can act on the appropriate network.
The different specific scenarios of unit stream are as follows: (1) static scenes, (2) moving objects, (3) a moving camera relative to static scenes, (4) a stationary camera relative to moving objects, and (5) a moving camera relative to a moving object. On the one hand, all scenarios provide depth cues as a dynamic supplement to the depth information of a single frame. On the other hand, only scenarios (1) and (3) provide camera pose information, which is an essential link in the process of view reconstruction, while (2), (4) and (5) are inappropriate since moving objects in them violate the underlying static scene assumption in view reconstruction. Due to lack of proper supervision signal, the depth estimation network comes to a deadlock in the area of dynamic objects. In addition, a single view cannot provide the dynamic properties of moving objects. PoseNet essentially learns motion information between frames. In order to make full use of inter-frame motion information, we first use the PoseNet encoder as the unit stream extraction network whose outputs are the unit stream to model the shallow space between frames, and then extract dynamic and static depth cues from complex and diverse implicit information in space. 

To better extract the implicit cues and estimate the depth in all above mentioned five scenarios, we innovatively propose two modules, i.e., IDCE and HAM based on the mainstream framework. The depth cues extracted by IDCE can be used as a dynamic supplement. The cues are modeled via statistics of convolutional activations, and perform an element-wise sum operation with the feature of the target frame, thereby increasing the proportion of moving objects features. And they can guide subsequent scene depth and make depth estimation near dynamic objects more accurate, while static proposals enforce the DepthNet to predict smooth depth changes over consecutive snippets. HAM can effectively reduce noise caused by moving objects in the cases of (2), (4) and (5) scenarios. The detailed architecture of our proposed framework is illustrated in Fig.~\ref{fig:overview}. 

\subsubsection{Implicit depth cues extractor}
It is shown in Fig.~\ref{fig:overview} (b) that the IDCE is an intermediate transition layer that links the stream encoding network and the DepthNet. It is designed to transfer implicit depth cues from the unit stream to the DepthNet. We adjust bottleneck~\cite{DBLP:conf/cvpr/HeZRS16} and use it as the basic block. Empirically, we cascade four identical bottlenecks as the final depth cues extractor. As shown in Fig.~\ref{fig:overview} (b), each bottleneck contains three layers, which are $1\times1$, $3\times3$, and $1\times1$ convolutions. The $1\times1$ layers are responsible for reducing the channel number to 1/4 and then restoring dimensions. All layers are performed with a stride of 1. Since the input and output are of the same dimensions, identity shortcuts are directly used.

\subsubsection{High-dimensional attention  module}

Attention can bias the allocation of available resources towards the most valuable parts of an input signal. Recently, the combination of spatial and channel attention module (CAM) has been successfully applied to a variety of vision tasks~\cite{DBLP:conf/cvpr/HuSS18,DBLP05819,DBLP:conf/cvpr/FuLT0BFL19,DBLP:journals/corr/abs-1910-08897}. Nonetheless, CAM cannot effectively reduce noise and is insufficiently rich to capture the high dimensional geometric characteristics of multiple views.

Inspired by~\cite{DBLP05819}, by extending features to high-dimensions using the Gaussian kernel, we propose a HAM as illustrated in Fig.~\ref{fig:overview} (c). Given a local feature $x\in \mathbb{R}^{C\times H \times W}$, where $C$, $H$ and $W$ denote channel, height and width dimensions, we first feed it into three independent $3\times 3$ convolution layers to generate three new features K, Q, V$\in \mathbb{R}^{C\times H \times W}$, respectively. After that, we perform a Gaussian kernel function between K and Q to find the similarity $s$ between each feature point K and Q. Uncertainties in the unit stream introduced by moving objects, occlusions, and incomplete Lambertian surfaces are controlled by the similarity $s$. Besides, it can search for global transformation in multiple views. Then we perform a matrix multiplication between $s$ and V. Finally we multiply it by a scale parameter $\beta$ and perform an element-wise sum operation with the feature $x$ to obtain global camera pose $x'\in \mathbb{R}^{C\times H \times W}$ as follows: 
\begin{align}
x' &= \beta \cdot s \cdot F_v(x)+x,
\label{fin_x}\\
\textrm{with} \ \ \ s &= e^{-\frac{\left\|F_k(x)-F_q(x)\right\|^2}{2\delta^2}},
\label{sim}
\end{align}
where $F_*(x)$ denotes the convolution layer with an activation function, $\delta$ is a hyperparameter, which is experimentally set to be 0.5. $\beta$ is initialized as 0 and gradually updated as the model learns.

Equation \eqref{sim} represents the approximate relationship between two tensors. Each element in Equation \eqref{sim} can be expanded into $n$th-order polynomial by Taylor's expansion:

\begin{align}
	e^{-\frac{\left|a-b\right|^2}{2\delta^2}}= e^{-\frac{a^2-2ab+b^2}{2\delta^2}}
	= \varphi (a)^T\varphi (b),
\end{align}
where
\begin{align}
\varphi (\theta) &= \begin{bmatrix}
e_0(\theta),
e_1(\theta),
e_2(\theta),
\cdots,
e_n(\theta),
\cdots
\end{bmatrix},
\label{phi}\\
e_n(\theta) &=  \sqrt{\frac{1}{n!}}\frac{\theta^n e^{-\frac{\theta^2}{2\delta^2}}}{\delta^n},n=0,1,2,....
\label{en_x}
\end{align}

In equation~\eqref{phi}, $\varphi(a)$ and $\varphi(b)$ are capable of modeling and using the high-order statistics of the local descriptor $a$ and $b$ ($a \in K, b \in Q$). Thus, we can directly obtain the high-order attention ‘map’ through equation~\eqref{sim}. The value is in the interval [0,1].  In equation~\eqref{en_x}, $e_n(\cdot)$ represents the component representation of the local descriptor in a n-dimensional space. Compared with the method that directly uses $k$ and $q$ to calculate the attention ‘map’, equation~\eqref{fin_x} comprehensively considers multi-dimensional similarity. When two tensors have similar components on each feature space, the tensors are globally similar. At the same time, this effectively suppresses the uncertainty of the local descriptor in the high-dimensional space. 

HAM can extract global dynamic transformation from the unit stream. The inter-frame features of the original space are mapped to the high-dimensional feature space, which captures more complex and high-order relationships, and matches the global spatial correlation of the original view.

\subsubsection{Network architecture}
By integrating the above mentioned two modules into the mainstream framework, we establish a new self-supervised depth estimation framework (see Fig.~\ref{fig:overview}). We rely on successful architecture in~\cite{DBLP:journals/corr/abs-1806-01260} as our basic framework. Both DepthNet encoder (DE) and PoseNet encoder (PE) use the same architecture (ResNet18~\cite{DBLP:conf/cvpr/HeZRS16}) except for the first layer. The first-level convolution channel of the PE is changed from 3 to 6, which allows the adjacent frames to feed into the network. PE is considered as a unit stream extraction network (USEN) and IDCE is used to connect the USEN and the DepthNet decoder (DD). The input size of IDCE is the same as the output of the USEN, and the output size is consistent with the input of the DD. We perform an element-wise sum operation at the last layer of IDCE and DE, then feed the results into DD. The DepthNet adopts a multi-scale architecture and predicts disparity maps with 1, 1/2, 1/4, 1/8 resolutions relative to the color image. HAM is used as the subsequent processing of PoseNet Decoder to obtain the final global 6D ego-motion. For the proposed modules, we adopt batch normalization right after each convolution and before ReLU activation.

\begin{table*}
	\centering
	\caption{Evaluation results of depth estimation on the KITTI improved ground truth~\cite{UhrigSSFBG17}.}

		\begin{tabular}{|c c|c c c c|c c c|}
			\hline
			
			Methods & Dataset & Abs Rel & Sq Rel &RMSE & RMSE log & {$\delta$ $ \textless$ 1.25 } & {$\delta$ $ \textless$ $1.25^2$} & {$\delta$ $ \textless$ $1.25^3$}\\
			\hline
			\hline
			
			Zhou et al.~\cite{8100183}  &M 			&0.176 &1.532& 6.129 &0.244 &0.758& 0.921& 0.971\\
             GeoNet~\cite{8578310} & M				&0.132 &0.994& 5.240 &0.193& 0.883& 0.953 &0.985\\
			Mahjourian et al.~\cite{DBLP:conf/cvpr/MahjourianWA18} & M	&0.134& 0.983& 5.501& 0.203& 0.827& 0.944 &0.981\\
            EPC++~\cite{abs-1810-06125}  & M	&0.120& 0.789& 4.755& 0.177& 0.856& 0.961 &0.987\\
		    Monodepth2~\cite{DBLP:journals/corr/abs-1806-01260} (192$\times$640) &  M &\underline{ 0.090}& \underline{0.545}&\underline{3.942}& \underline{0.137}&\underline{ 0.914}& \underline{0.983}& \underline{0.995}\\
			\textbf {Ours} (192$\times$640) &  M&\textbf{ 0.082}& \textbf{0.462}& \textbf{3.739}&\textbf {0.127}& \textbf{0.923}&\textbf{0.984}&\textbf {0.996}\\
			\hline
			EPC++~\cite{abs-1810-06125} & MS	&0.123& 0.754& 4.453& 0.172& 0.863& 0.964 &0.989\\
		    Monodepth2~\cite{DBLP:journals/corr/abs-1806-01260} (192$\times$640) &  MS 			&\underline{ 0.080}& \underline{0.466}& \underline{3.681}& \underline{0.127}&\underline{ 0.926}& \underline{0.985}& \underline{0.995}\\
			\textbf {Ours} (192$\times$640) &MS				& \textbf{0.077}&  \textbf{0.431}& \textbf{3.598}& \textbf{0.121}& \textbf{0.931}& \textbf{0.986}& \textbf{0.996} \\
			
			\hline
	\end{tabular}
	
	\label{table:improved}
\end{table*}

\section{Experiments}

\subsection{Dataset}

Our experiments are mainly conducted on KITTI \cite{DBLP:journals/ijrr/GeigerLSU13}, CityScapes \cite{7780719} and Make3D \cite{4531745} datasets. The KITTI dataset includes a full suite of raw data such as stereo videos and 3D point clouds. We use 39810 monocular frames and stereo pairs for training, about 4K images for evaluation, and 697 images from the test split~\cite{DBLP:conf/nips/EigenPF14}. The CityScapes dataset contains various stereo video sequences recorded from 50 different cities. We choose the monocular sequence of the 8-bit image taken by the left monocular camera, and additionally evaluate our model trained by KITTI on Make3D dataset, which is unseen during training to evaluate the generalization ability. Also, we pre-train the network on CityScapes and finetune on KITTI. 

As for the experimental metrics, following Zhou et al.~\cite{8100183}, we use the following metrics to evaluate our depth estimation method on the KITTI test split and Make3D dataset: (1) Abs Rel, Sq Rel, RMSE and log RMSE (lower the better), and (2) $\delta$ $ \textless$ 1.25, $\delta$ $ \textless$ $1.25^2$, $\delta$ $ \textless$ $1.25^3$ (higher the better).

 The median scaling~\cite{8100183} is used to align the predictions with the ground truth during the evaluation. Note that we remove the sequences where the camera does not move between frames during training. During the evaluation, two adjacent frames ($I_{t-1}$, $I_t$) are fed to USEN and DE. For discrete samples, such as the first frame of a video, we duplicate each sample to simulate adjacent frames.

\subsection{Implementation details}

Our model is implemented with the PyTorch~\cite{pytorch} framework and a single Tesla V100, trained for 20 epochs, with a batch size of 8. Additionally, random contrast, brightness, saturation, color jittering, horizontal flip, random resizing are used during training. The default input and output resolution is 192$\times$640. At the same time, for comparsions, we also use a larger resolution 320$\times$1024 in experiments. 

Similar to~\cite{DBLP:journals/corr/abs-1806-01260,DBLP:conf/nips/BianLWZSC019}, the DE and USEN are initialized by a ResNet-18 backbone pretrained on the ImageNet dataset~\cite{DBLP:journals/ijcv/RussakovskyDSKS15}. USEN uses the pre-training weights and removes the weights of the first layer. We adopt Adam~\cite{DBLP:journals/corr/KingmaB14} optimizer with an initial learning rate of $1e$-$4$, and reduce it to 10\% after 15 epochs. $\beta_1$, $\beta_2$ and weight decay are set to be 0.9, 0.999 and 0.0001 respectively. In order to alleviate the difficulty of directly optimizing the IDCE and HAM, an effective training strategy is explored to decouple the disparity images from the transformation. More precisely, we first train the baseline and HAM, then jointly train the entire model. It turns out that this strategy leads to superior performances on multiple datasets.

\subsection{Comparisons with the SOTA}

In this subsection, our methods are evaluated from both qualitative and quantitative point of views on the KITTI, the Make3D datasets, and further evaluate odometry results on KITTI odometry dataset. Results show that our proposed framework achieves SOTA performance, and outperforms recent algorithms on the depth estimation tasks.

\begin{figure}
\centering
\includegraphics[width=8cm]{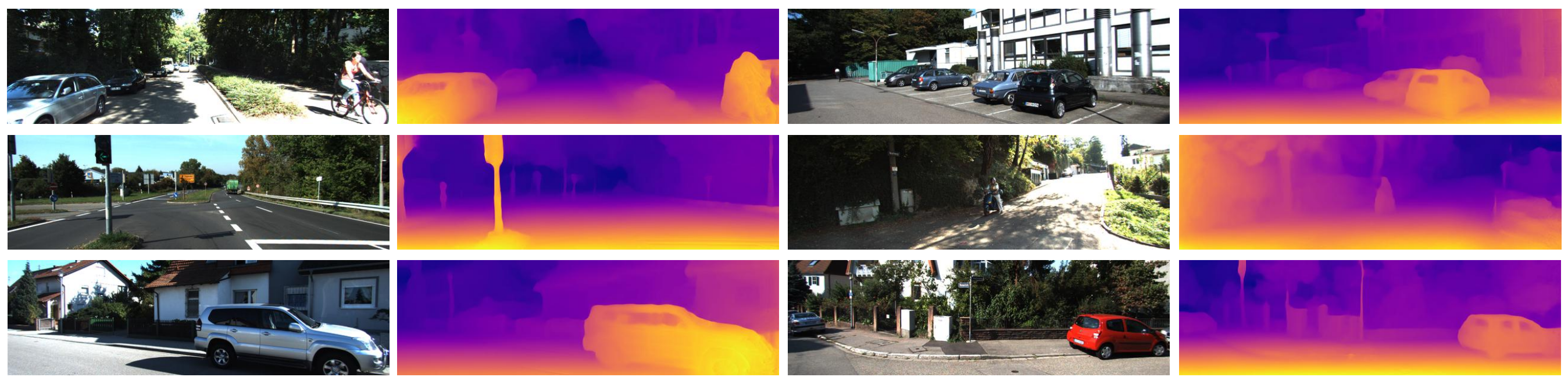}
\caption{More qualitative results on KITTI test splits.}
\label{fig:results2}
\end{figure}

\subsubsection{Results on KITTI dataset}

We compare the performance of the proposed framework with the baseline, as well as existing SOTA methods as shown in Table~\ref{table:T1}.
Results show that our method achieves significant gains over all existing SOTA self-supervised approaches when trained with different types datasets, which are KITTI monocular frames only, KITTI monocular frames and stereo pairs, CityScapes and KITTI monocular frames.

We summarize the main results in Table~\ref{table:T1} as follows: (1) Overall, our method outperforms previous SOTA on the same training setting. Although trained in a self-supervised manner, our method competes quite favorably with most supervised baselines. (2) It is observed that our KITTI monocular and Cytiscapes KITTI model results are slightly lower than~\cite{bozorgtabar2019syndemo} on the Sq Rel and RMSE metrics, and the high-resolution monocular-stereo pairs model obtains a second performance 0.101 on Abs Rel, only 0.001 less than the result in~\cite{watson2019self}. However, it should be mentioned that~\cite{bozorgtabar2019syndemo,watson2019self} use a new auxiliary supervision signal while we only use Cytiscapes and KITTI raw data. (3) For KITTI monocular training, our method is slightly better than~\cite{DBLP:journals/corr/abs-2001-02613} which is trained by a ConvLSTM-based network with video inputs. Compared with them, we improve the mainstream framework and our method is much simpler and more efficient. (4) It is worth mentioning that our method outperforms recent work~\cite{DBLP:journals/corr/abs-1907-05820,8953778,8578310,DBLP:conf/eccv/ZouLH18} that jointly learns multiple tasks as well as complex network structure. (5) Moreover, experimental results that the stereo view, CityScapes pre-training, and high-resolution images can improve the performance of the monocular depth estimation model. 

As shown in Table~\ref{table:improved}, we directly compare the proposed method with existing methods on the KITTI improved ground truth from~\cite{UhrigSSFBG17}. The imporoved depth provides 652 of the 697 test frames contained in~\cite{EigenF15}. The predicted depth maps are clipped to 80 meters, and then the full maps are evaluated. The values of the existing methods are reported by~\cite{DBLP:journals/corr/abs-1806-01260}. Our method is still significantly better than existing published methods without retraining.

Qualitative results are shown in Fig.~\ref{fig:results}, where some comparison samples between our KITTI monocular method and some self-supervised baselines are presented. As shown in the first image, compared with other methods, our method provides a clearer motion contour. It also perceives the geometry of static objects and results in a more reasonable depth estimation. Moreover, the depth difference between static overlapping objects can be distinguished significantly. In order to comprehensively visualize the performance of the proposed method, more qualitative results in different cases on the KITTI dataset are shown in Fig.~\ref{fig:results2}.

\begin{table}
	\centering
	\caption{Evaluation of depth estimation results on the Make3D test set\cite{4531745}. The best results in each category are in bold, and the second best are underlined. Full denotes the proposed self-supervised joint learning framework.}
	
	\begin{tabular}{|c c|c c c|}
		\hline
		
		Method & Train  & Abs Rel & Sq Rel &RMSE \\
		\hline
		\hline
		Zhou et al.~\cite{8100183}  		& M & 0.383 & 5.321 & 10.470  \\
		DDVO~\cite{8099721} 		& M & 0.387 & 4.720 & 8.090\\
		Monodepth2~\cite{DBLP:journals/corr/abs-1806-01260}	& M & 0.322 & 3.589 &7.417\\
		SynDeMo~\cite{bozorgtabar2019syndemo}  	& M & 0.330 & 2.692 & 6.850 \\
		Bian et al.~\cite{DBLP:conf/nips/BianLWZSC019}	  	& M & \underline{0.312}	&  $-$    &  $ - $ \\
		Zhou et al.~\cite{DBLP:journals/corr/abs-1910-08897}		& M & 0.318 & 2.288 &\textbf{ 6.669}\\
		\hline
		Ours	 &M&\textbf{ 0.306}  & \textbf{  2.056}  &    \underline{ 6.721} \\

		\hline
	\end{tabular}
	
	\label{table:T2}
\end{table}

\subsubsection{Results on Make3D dataset}
In Table \ref{table:T2}, we directly evaluate our method's performance on Make3D dataset without any training data on it. Our model is trained on KITTI monocular video without any fine-tuning. Following the evaluation protocol in~\cite{8100183}, only using central images where depth is less than 70 meters are evaluated. Our result outperforms existing SOTA methods that do not use depth supervision, showing excellent cross-dataset generalization ability. 

	\begin{figure}
		\centering
		\includegraphics[width=8cm]{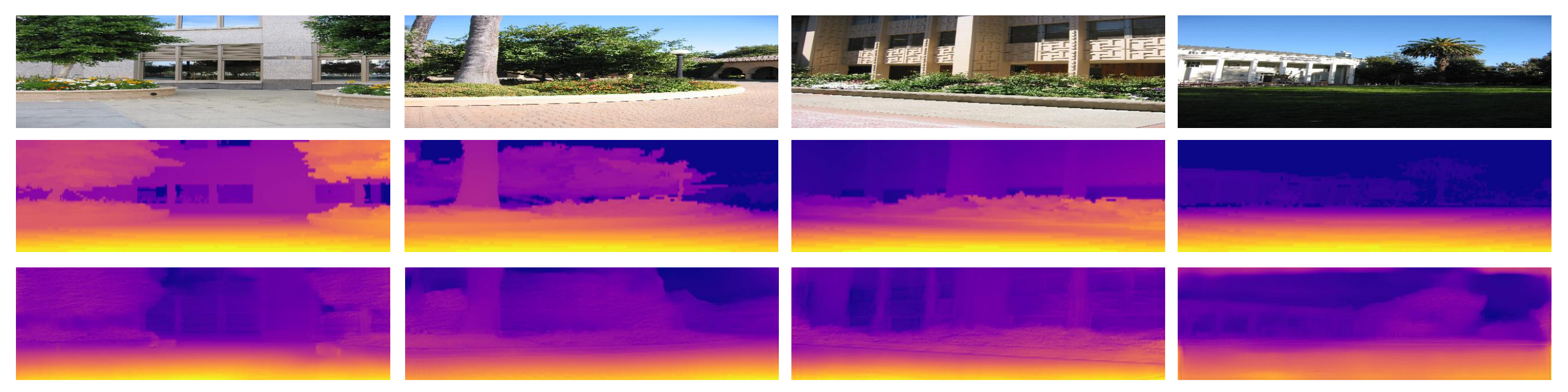}
		\caption{Qualitative results on Make3D dataset. From top to bottom, the images are input, ground truth, and results of our CS+K monocular method, respectively.}
		\label{fig:make3d}
	\end{figure}

Indeed, our method cannot be directly applied on the Make3D dataset because the data is discrete and does not have the inherent dynamic properties of video sequences. In order to adapt our method to this dataset, we resize the test image to 192$\times$640 resolution and replicate each sample to simulate adjacent frames for evaluation. The results clearly demonstrate the existence of static depth cues in shallow space are helpful for depth estimation.

Fig.~\ref{fig:make3d} shows some qualitative results of the Make3D dataset, which are estimated by our CS+K model. Both quantitative and qualitative experiments demonstrate the generalization ability of our method in estimating accurate depth maps from consecutive frames.

\begin{table}
	\centering
	\caption{ Evaluation results on the KITTI odometry dataset. `\#frames' is the number of input frames.}

	\begin{tabular}{|c |c |c| c|}
		\hline
		      -           &  Sequence 09 &Sequence 10 &\#frames\\
	    
		\hline
        \hline
		ORB-SLAM (full) &  0.014$\pm$0.008 &0.012$\pm$0.011 &-\\
		\hline

		Zhou et al.~\cite{8100183} & 0.021$\pm$0.017 	&  0.020$\pm$0.015   &5\\  
		SynDeMo~\cite{bozorgtabar2019syndemo}& 0.011$\pm$0.007 & 0.011$\pm$0.015 &5\\
		Monodepth2~\cite{DBLP:journals/corr/abs-1806-01260} &0.017$\pm$0.008 & 0.015$\pm$0.010 &2\\
		Zhou et al.~\cite{DBLP:journals/corr/abs-1910-08897} &0.015$\pm$0.007& 0.015$\pm$0.009 &3\\
        GLNet~\cite{DBLP:journals/corr/abs-1907-05820} &\textbf{0.011$\pm$0.006} &\textbf{0.011$\pm$0.009} &3\\
\hline
 		
		Ours &0.016$\pm$0.008 	&0.014$\pm$0.009	  &2\\
       \hline
	\end{tabular}

	\label{table:Todom}
\end{table}

\subsubsection{Results on KITTI odometry dataset}
 For completeness, we evaluate the two-frame model on a five-frame test sequence and combine four frame-to-frame transformation in each group to form a local trajectory. We measure the absolute trajectory error averaged over every 5-frame snippets on sequences 9 and 10. The pose estimation results are summarized in Table~\ref{table:Todom}. Although our method does not exceed SOTA, still reamains a satisfied performance. The main advantage of our method is reflected in the depth estimation task. 

\subsection{Ablation study}

To analyze individual effects of each component in our framework, we first perform ablation studies on the KITTI and CityScapes by replacing various components. Then our modules are applied to other methods to evaluate its generalization ability. Finally, we experiment on images with different resolutions.

\begin{table*}
	\centering
	\caption{Evaluation of each component in our framework on KITTI’s eigen test split. M and S denote monocular video and stereo supervision. K and CS are KITTI and CityScapes datasets.}

	\begin{tabular}{|c c|c c c c|c c c|}
		\hline
	
        Train & Methods & Abs Rel & Sq Rel &RMSE & RMSE log & {$\delta$ $ \textless$ 1.25 } & {$\delta$ $ \textless$ $1.25^2$} & {$\delta$ $ \textless$ $1.25^3$}\\
		\hline
        \hline

		K (M) & Baseline						& 0.121  &   0.899  &   4.934  &   0.199  &   0.856  &   0.955  &   0.980  \\
 		K (M) & Baseline+IDCE					& 0.117  &   0.875  &   4.829  &   0.196  &   0.862  &   0.956  &   0.981  \\
		K (M) & Baseline+HAM 		&0.112  &   0.855  &   4.781  &   0.190  &   0.878  &   0.960  &   0.981   \\
		K (M) & Baseline+HAM+IDCE	&{ \textbf{0.106}}& {\textbf{0.799}}& {\textbf{4.662}}& {\textbf{0.187}}& {\textbf{0.889}}&{\textbf{0.961}}& {\textbf{0.982}}\\
        \hline
		K (MS) &Baseline						& 0.114  &   0.897  &   4.837  &   0.193  &   0.877  &   0.959  &   0.981 \\
		K (MS) &Baseline+IDCE					& 0.108  &   0.817  &   4.677  &   0.189  &   0.884  &   0.960  &   0.981  \\
		K (MS) &Baseline+HAM			& 0.107  &   0.816  &   4.663  &   0.187  &   0.887  &   0.961  &   0.981  \\ 
		K (MS) &Baseline+HAM+IDCE	&{\textbf{0.102}}&{ \textbf{0.776}}&{ \textbf{4.534}}& {\textbf{0.183}}&{ \textbf{0.893}}& {\textbf{0.963}}& {\textbf{0.982}} \\ 
       \hline
 		CS (M) &Baseline						&   0.194  &   1.340  &   5.896  &   0.256  &   0.697  &   0.919  &   0.974 \\
		CS (M) &Baseline+HAM			&0.189  &   1.354  &   5.859  &   0.253  &   0.706  &   0.922  &   0.974 \\
		CS (M) &Baseline+HAM+IDCE	&\textbf{0.169}  &   \textbf{1.303}  &   \textbf{5.706}  &   \textbf{0.238}  &   \textbf{0.765}  &   \textbf{0.933}  &   \textbf{0.974}  \\
       \hline
	\end{tabular}

	\label{table:T3}
\end{table*}
For ablation studies, we use the baseline~\cite{DBLP:journals/corr/abs-1806-01260} and images of 192$\times$640 resolution. As shown in Table~\ref{table:T3}, results demonstrate that the proposed modules provide benefits in different perspectives. Compared with IDCE, HAM achieves better performance. We hypothesize that the noise in the unit stream has a great impact on the self-supervised depth estimation task, and the global transformation obtained by HAM can effectively reduce uncertainty. To verify this point, we perform statistics on the number and channel dimensions of the HAM processed feature. As shown in Fig.~\ref{fig:ham}, the visual plane becomes smoother after the processing of HAM. On this basis, we add the IDCE module to transfer implicit depth cues from the unit stream to the DepthNet. The depth cues can be used as a dynamic supplement for DepthNet. When combined them together, our proposed method achieves SOTA results.

\begin{small}
	\begin{figure}
		\centering
		\includegraphics[width=8cm]{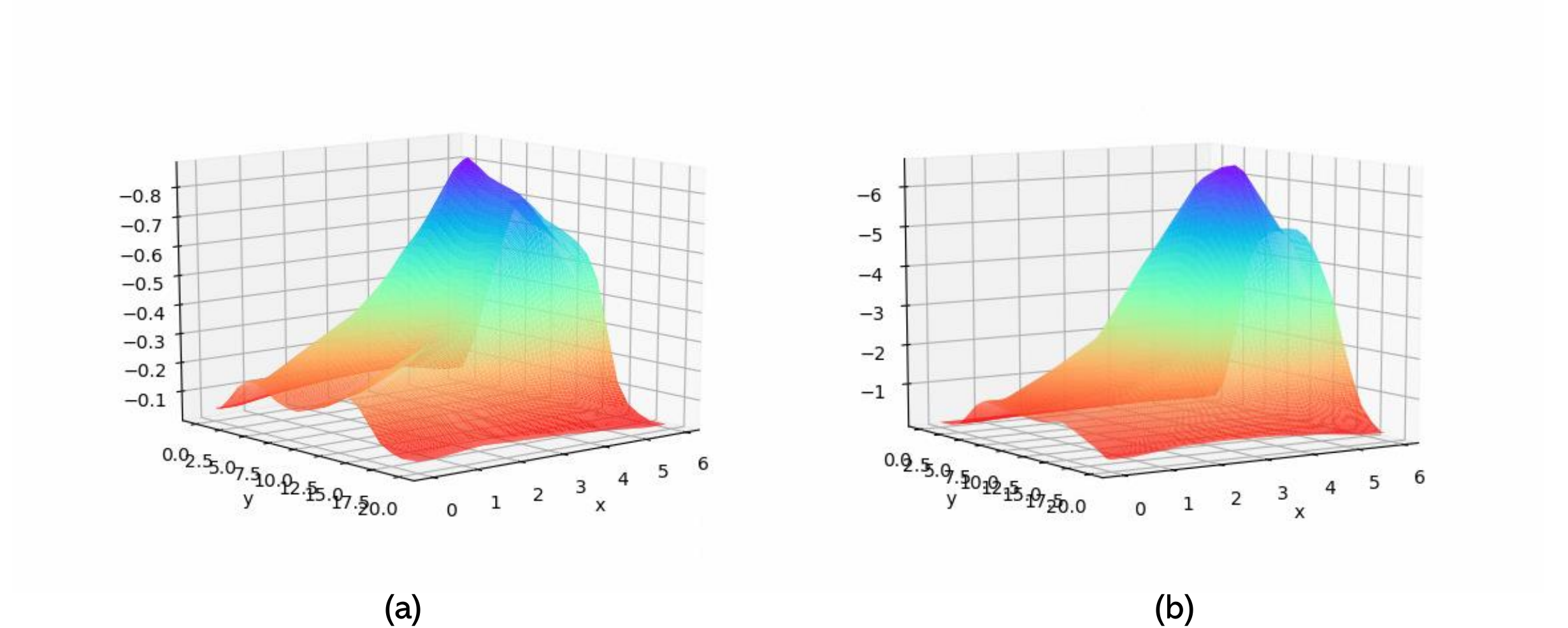}
		\caption{Visualization results of PoseNet feature statistics. (a) is feature statistics without the post-processing by HAM. (b) is feature statistics with the post-processing by HAM.}
		\label{fig:ham}
	\end{figure}
\end{small}

\begin{figure}
		\centering
		\includegraphics[width=9cm]{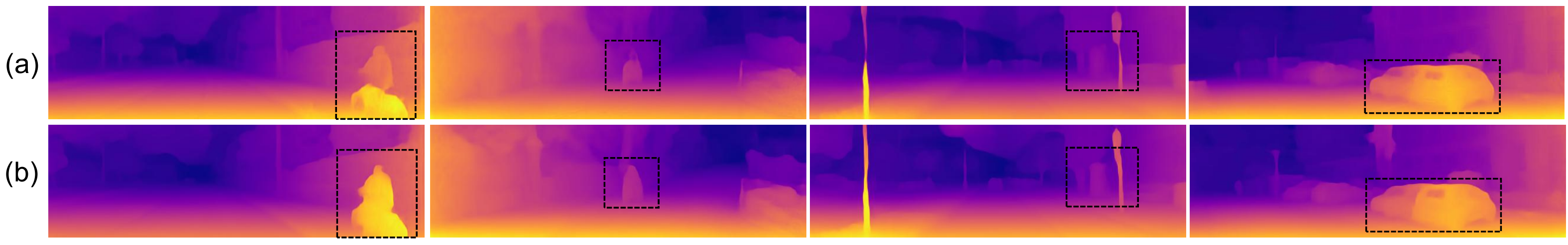}
		\caption{Comparision of the depth map at the object boundary. (a) Predicted depth maps without IDCE. (b) Predicted depth maps with IDCE.}
		\label{fig:idce}
	\end{figure}

Fig.~\ref{fig:idce} shows the comparision of the depth map at the object boundary, including two dynamic scenes and two static scenes. Compared with baseline, it can be seen that IDCE can effectively reduce the motion blur phenomenon by combining the depth information between adjacent frames, and can provide a clearer contour for dynamic or static objects.

\begin{table*}
	\centering
	\caption{Evaluation of different attention modules on different PoseNets.}

	\begin{tabular}{|c cc|c c c c|c c c|}
		\hline
	    Pose network & Attention Module &Train &  Abs Rel & Sq Rel &RMSE & RMSE log & {$\delta$ $ \textless$ 1.25 } & {$\delta$ $ \textless$ $1.25^2$} & {$\delta$ $ \textless$ $1.25^3$}\\
     
		\hline
        \hline
		ResPose CNN & -  	&  MS  	& 0.114  &   0.897  &   4.837  &   0.193  &   0.877  &   0.959 &0.981\\  
		ResPose CNN & CAM    &  MS   &0.108  &   0.853  &   4.782  &   0.189  &   0.886  &   0.960  &   0.981  \\
		ResPose CNN & HAM 	&  MS  	& \textbf{0.107}  &   \textbf{0.816}  &   \textbf{4.663}  &   \textbf{0.187}  &   \textbf{0.887}  &   \textbf{0.961}  &   \textbf{0.981}  \\ 
        \hline
 		Pose CNN & -		&M		&0.136  &   1.067  &   5.274  &   0.209  &   0.840  &   0.950  &   0.979  \\
		Pose CNN & CAM	&M		&0.139  &   1.080  &   5.326  &   0.211  &   0.831  &   0.949  &   0.979  \\
		Pose CNN & HAM	&M		&\textbf{0.136}  &   \textbf{1.024}  &   \textbf{5.151}  &   \textbf{0.207}  &   \textbf{0.849}  &   \textbf{0.952}  &   \textbf{0.980}  \\
       \hline
	\end{tabular}

	\label{table:T4}
\end{table*}

\begin{table*}
	\centering
	\caption{Ablation studies on different image resolutions.}

	\begin{tabular}{|c c|c c c c|c c c|}
		\hline
	
        Train & Resolution & Abs Rel & Sq Rel &RMSE & RMSE log & {$\delta$ $ \textless$ 1.25 } & {$\delta$ $ \textless$ $1.25^2$} & {$\delta$ $ \textless$ $1.25^3$}\\
		\hline
        \hline
		K (M) & (128$\times$416)	&   0.113  &   0.864  &   4.872  &   0.194  &   0.872  &   0.956  &   0.980  \\
		K (M) & (192$\times$640)	&{ 0.106}& {0.799}& {4.662}& {0.187}& {0.889}&{0.961}& {0.982}\\
		K (M) & (320$\times$1024)	&\textbf{ 0.106}& \textbf{0.773}& \textbf{4.491}&\textbf {0.185}& \textbf{0.890}&\textbf{0.962}&\textbf {0.982}\\
        \hline

		K (MS) &(128$\times$416)&{0.106}&{ 0.774}&{ 4.623}& {0.184}&{ 0.886}& {0.962}& {0.983} \\ 
		K (MS) &(192$\times$640)&{0.102}&{ 0.776}&{ 4.534}& {0.183}&{ 0.893}& {0.963}& {0.982} \\ 
		K (MS) &(320$\times$1024)&\textbf{0.101}&\textbf{ 0.725}&\textbf{ 4.360}& \textbf{0.179}&\textbf{ 0.898}& \textbf{0.965}& \textbf{0.983} \\ 
  		
       \hline
 		CS+K (M) &(128$\times$416)& 0.113  &   0.866  &   4.843  &   0.195  &   0.874  &   0.955  &   0.980  \\
		CS+K (M) &(192$\times$640)&{0.106}&{ 0.774}&{ 4.623}& {0.184}&{ 0.886}& {0.962}& \textbf{0.983} \\
		CS+K (M) &(320$\times$1024)&\textbf{0.104}&\textbf{ 0.771}&\textbf{ 4.463}& \textbf{0.183}&\textbf{ 0.893}& \textbf{0.963}& {0.982} \\

       \hline
	\end{tabular}

	\label{table:T5}
\end{table*}

The generalization ability of HAM is evaluated by adopting different PoseNets. It is shown in Table~\ref{table:T4} the benefit that our HAM brings to ResPose CNN~\cite{DBLP:journals/corr/abs-1806-01260} and Pose CNN~\cite{8100183}, while the basic attention module such as CAM gains negative yields on Pose CNN. We conjecture that this is due to the noisy unit-stream space provided by the Pose CNN. The basic attention is not conducive to capturing reasonable geometric transformations from sophisticated shallow space, while the HAM produces more robust results. 

Table~\ref{table:T5} shows the results in different training settings on images with different resolutions. It shows that high-resolution images improve performance but increase training time. It takes approximately 9 hours for training the K(M) (128$\times$416) model, while (320$\times$1024) model takes about 49 hours.

\section{Conclusion}
To solve the depth discontinuity and motion artifact problems, a novel self-supervised joint learning framework is proposed. Our main idea is to take advantage of the unit stream, which represents the spatial and temporal information in consecutive frames. The proposed framework utilizes implicit cues extractor to extract static and dynamic depth cues from unit stream in shallow space, and uses implicit cues to guide the depth estimation of a single image. Moreover, a high-dimensional attention module is introuced to extract global pose information, effectively reducing appearance loss. Extensive experimental results demonstrate that our method outperforms SOTA performance on the KITTI/Make3D dataset by a significant margin, and this framework can be generalized to any self-supervised monocular depth estimation network. For the future work, it is worthwhile to explore a more accurate visual odometry based on this framework.

%
%
\bibliographystyle{splncs04}
\bibliography{main}
\end{document}